\documentclass[sigconf,anonymous=false]{acmart}

\renewcommand\footnotetextcopyrightpermission[1]{} 

\AtBeginDocument{%
  \providecommand\BibTeX{{%
    \normalfont B\kern-0.5em{\scshape i\kern-0.25em b}\kern-0.8em\TeX}}}

\acmConference[TAHRI]{Make sure to enter the correct
  conference title from your rights confirmation emai}{March 9–10,
  2024}{Boulder, CO}
\acmBooktitle{Technological Advances in Human-Robot Interaction,
 March 9–10, 2024, Boulder, CO, USA} 

\usepackage{lipsum}

\usepackage{pifont}
\newcommand{\cmark}{\ding{51}}%
\newcommand{\xmark}{\ding{55}}%

\usepackage{algorithm}
\usepackage[noend]{algpseudocode}
\algrenewcommand\algorithmicrequire{\textbf{Input:}}
\algrenewcommand\algorithmicensure{\textbf{Output:}}
\algrenewcommand\algorithmicindent{0.5em}%

\algdef{SE}[SUBALG]{Indent}{EndIndent}{}{\algorithmicend\ }%
\algtext*{Indent}
\algtext*{EndIndent}

\usepackage{multirow}
\usepackage{array}
\newcolumntype{P}[1]{>{\centering\arraybackslash}p{#1}}
\usepackage[font=small,skip=0pt]{caption}

\begin{document}

\makeatletter
\let\ACM@origbaselinestretch\baselinestretch

\title{O-TALC: Steps Towards Combating Oversegmentation within Online Action Segmentation}

\author{Matthew Kent Myers}
\authornote{Primary Author}
\email{M.J.Kent-Myers2@ncl.ac.uk}
\affiliation{%
  \institution{School of Engineering}
  \city{Newcastle University}
  \country{UK}
}

\author{Nick Wright}
\affiliation{%
  \institution{School of Engineering}
  \city{Newcastle University}
  \country{UK}
}

\author{A. Stephen McGough}
\affiliation{%
  \institution{School of Computing}
  \city{Newcastle University}
  \country{UK}
}

\author{Nicholas Martin}
\affiliation{%
  \institution{Tharsus Ltd.}
  \city{Blyth}
  \country{UK}
}

\renewcommand{\shortauthors}{Kent Myers and Wright, et al.}

\begin{abstract}
Online temporal action segmentation shows a strong potential to facilitate many HRI tasks where extended human action sequences must be tracked and understood in real time.  Traditional action segmentation approaches, however, operate in an offline two stage approach, relying on computationally expensive video wide features for segmentation, rendering them unsuitable for online HRI applications. In order to facilitate online action segmentation on a stream of incoming video data, we introduce two methods for improved training and inference of backbone action recognition models, allowing them to be deployed directly for online frame level classification. Firstly, we introduce surround dense sampling whilst training to facilitate training \textit{vs.} inference clip matching and improve segment boundary predictions. Secondly, we introduce an \textbf{O}nline \textbf{T}emporally \textbf{A}ware \textbf{L}abel \textbf{C}leaning (\textit{O-TALC}) strategy to explicitly reduce oversegmentation during online inference. As our methods are backbone invariant, they can be deployed with computationally efficient spatio-temporal action recognition models capable of operating in real time with a small segmentation latency. We show our method outperforms similar online action segmentation work as well as matches the performance of many offline models with access to full temporal resolution when operating on challenging fine-grained datasets. 
\end{abstract}






\maketitle

\section{Introduction \& Background}

\begin{figure}
    \centering
    \includegraphics[width=0.48\textwidth]{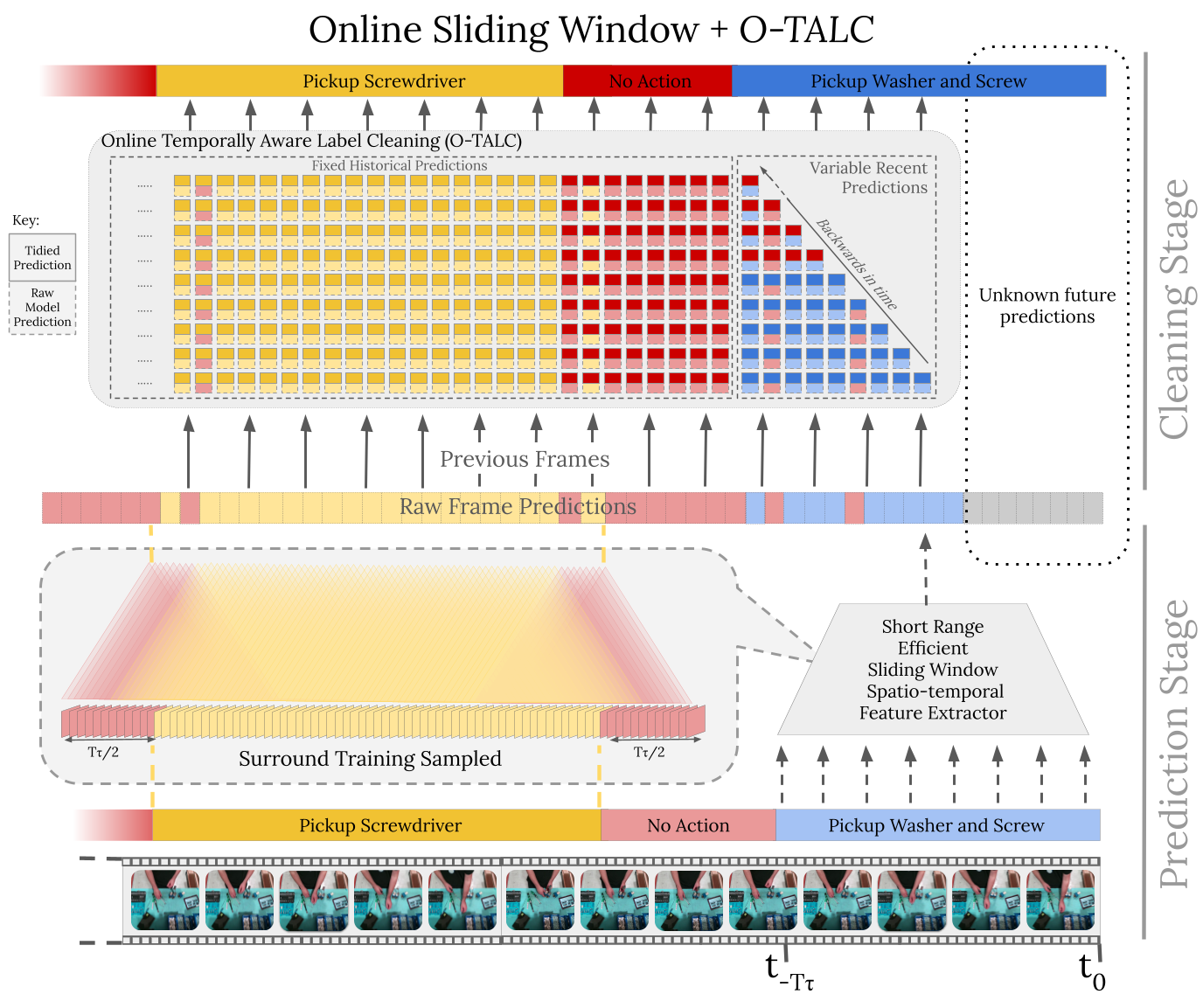}
    \caption{\small An overview of our outlined online action segmentation approach utilising a surround sampled trained backbone model operating in a sliding window fashion with \textit{O-TALC.}}
    \label{fig:overview}
    \vspace{-10pt}
\end{figure}

Video data forms a key modality for machines to perceive the world around them, with many human robot interaction (HRI) scenarios, especially those within manufacturing, requiring the real time understanding of human actions in order to perform collaboration, assistance, analysis or guidance \cite{HA-VID,notification,weakly}. For tasks requiring an understanding of extended human action sequences, online temporal action segmentation (AS) forms an attractive method for the tracking and understanding of humans through the use of atomic action \cite{hands,HAA} analysis via graphs \cite{mistake} or formal task-models \cite{tracking}.

Offline temporal AS, formally defined as producing frame wise classification predictions $y_{1:N} = (y_1,y_2,..., y_N)$ from a given input video sequence containing N frames $\mathcal{X} = (x_1,x_2,..., x_N)$ where $y_i$ belongs to a set of known domain specific classes $\mathcal{C}$ has received much interest \cite{AS_lit}, with state-of-the-art models operating in two stages. First, a strong backbone extractor is utilised in a sliding window fashion to produce frame wise features, with a secondary long term temporal classification step deploying methods such as temporal convolutions \cite{TCN,ms-TCN,C2F-TCN}, transformers \cite{long-context}, or stable diffusion \cite{diff-AS} to produce refined segmentation predictions. Whilst these two stage models provide strong performances, the requirement for full video sequences at inference renders them unsuitable for applications requiring real time HRI. 

On the other hand, the task of online AS, whereby models must perform frame wise classification in real time (with some defined minimal classification delay) only utilising previously observed video frames, has received relatively little exploration. Current approaches have followed the more common online action detection approach of deploying backbone action recognition models in a sliding window fashion e.g \cite{woad,Movie-net,TSM,OnlineAD,cooking,LapNet,RNN} for frame level predictions. However, due to their focus on rapid action classification over accurate segmentation, these models often lead to poor boundary predictions and significant oversegmentation \cite{Efficient}. 

In order to combat these problems the work within this paper tackles the challenge of extending efficient action recognition backbone models to the online AS domain through the introduction of two backbone invariant improvements when operating models in a sliding window paradigm. We show that through the use of surround sampling whilst training, along with a novel, yet simple, \textbf{O}nline \textbf{T}emporally \textbf{A}ware \textbf{L}abel \textbf{C}leaning (\textit{O-TALC}) strategy targeting accurate segmentation over rapid detection, a strong online AS performance, rivalling offline approaches, can be achieved whilst operating in real time with only a modest segmentation delay of $\sim$2-5 seconds.

\section{Methodology}

\begin{table}[]
\small
\begin{tabular}{l|c|c} \toprule
\begin{tabular}[c]{@{}l@{}}\textbf{Training} \\ \textbf{Sampling} \\ \textbf{Strategy}\end{tabular} & \textbf{Initial Frame Selection}  $s.t. \ \ x_{-T\tau} \in \mathbb{N}$ & \begin{tabular}[c]{@{}l@{}}\textbf{AS label} \\ \textbf{Utilisation}\end{tabular} \\ \midrule
\begin{tabular}[c]{@{}l@{}}Dense \\ e.g I3D \cite{i3d} \end{tabular} & $x_{-T\tau}$ = ${\scriptsize \begin{cases} U \{N_s,N_e-T\tau\} & \text{if} \ (N_e - N_s) > T\tau \\ N_s & \text{otherwise}. \end{cases} }$ & \xmark \\ \hline
\begin{tabular}[c]{@{}l@{}}Surround \\ (ours)\end{tabular} & $x_{-T\tau} = U\{N_s-\frac{T\tau}{2} , N_e-\frac{T\tau}{2}\}$ & \cmark \\ \bottomrule
\end{tabular}
\caption{Initial frame index selection during training for traditional dense and surround dense sampling.}\label{tab:sample}
\vspace{-15pt}
\end{table}

The following section introduces our approach to extending short range spatio-temporal action recognition models to the online AS domain via a sliding window paradigm operating on inference clips\footnote{This introduces an online delay of $\sim$2 seconds, however significantly improves classification on spatially similar atomic actions \cite{hands}.} created by sampling T (8 or 16) previous frames at a temporal stride of $\tau$ (6 or 8), with an initial frame at -T$\tau$. Our method breaks down into two backbone invariant contributions, summarised in Fig. \ref{fig:overview}. We first introduce surround sampling in section \ref{sec:surround} to replace the traditional dense sampling approach commonly adopted whilst training action recognition models in order to ensure clip matching between training and online inference. We then introduce a \textit{O-TALC} in section \ref{sec:talc} for combating oversegmentation in an online manner.

\subsection{Surround Sampling During Training} \label{sec:surround}

\begin{algorithm}
    \caption{\small Real time Online Temporally Aware Label Cleaning ($\textit{O-TALC}$) algorithm for tidying a set of frame predictions up to the current frame t. Capable of running at over 1000 FPS on consumer CPUs. }\label{alg:TALC}
\begin{algorithmic}[5]
    \small
    \Require  $\hat{y}_{t}$, $\hat{y}_{:t-1}$, $\hat{y}^{tidy}_{:t-1}$, $b \in \mathbb{N}^1 < C_{min}$
    \Ensure $\hat{y}_{:t}^{tidy}$

    \State if $\hat{y}_{:t-1}$ empty:
        \Indent \State $\hat{y}^{tidy}_{:t}$ $\gets$ [$\hat{y}^{tidy}_{:t-1}$,$\hat{y}_{t}$]
    \EndIndent
    \State else:
        \Indent \State if $\hat{y}_{t}$ != $\hat{y}_{:t-1}$: \Comment{\scriptsize If there is a change in segment \small}
            \Indent \State $\hat{y}^{tidy}_{:t}$ $\gets$ [$\hat{y}^{tidy}_{:t-1}$,$\hat{y}^{tidy}_{t-1}$] \Comment{\scriptsize Segment only just changed, append previous class. \small}
        \EndIndent
        \State else:
            \Indent \State $have\_raw\_seg\_length$ $\gets$ $False$
            \State for $i$ in backwards loop: \Comment{\scriptsize Get length of segment via looping backwards \small}
                \Indent \State if $i == 0:$ \Comment{\scriptsize If beginning of sequence is reached (first segment) \small}
                    \Indent \State $\hat{y}^{tidy}_{:t}$ $\gets$ [$\hat{y}^{tidy}_{:t-1}$, $\hat{y}_{0}$] \Comment{\scriptsize Append the first segment prediction \small}
                            \State $\textbf{break}$
                    \EndIndent \State if $\hat{y}_{i}$ $\neq$ $\hat{y}_{i-1}$: \Comment{\scriptsize If there is a change in prediction \small}
                        \Indent \State if not $have\_raw\_seg\_length$:
                            \Indent \State $\texttt{raw\_seg\_length}$ $\gets$ t - i + 1
                            \State $have\_raw\_seg\_length$ $\gets$ $True$
                        \EndIndent \State if $\hat{y}_{i}$ not in $\hat{y}_{i-b:i}$:  \Comment{\scriptsize Secondary check allowing oversegmentation \small}
                            \Indent \State $\texttt{full\_seg\_length}$ $\gets$ t - i + 1
                                    \State if $\texttt{full\_seg\_length}$ $\ge C_{min}$ and $\texttt{raw\_seg\_length}$ $\ge b$:
                                        \Indent \State $\hat{y}^{tidy}_{i:t-1} \gets \hat{y}_{t}$ \Comment{\scriptsize Replace all segment frame predictions at once \small}
                                                \State $\hat{y}^{tidy}_{:t}$ $\gets$ [$\hat{y}^{tidy}_{:t-1}$, $\hat{y}_{t}$] \Comment{\scriptsize Append new valid frame prediction\small}
                                                \State $\textbf{break}$
                                            \EndIndent \State else: \Comment{\scriptsize If not statistically valid\small}
                                                \Indent \State $\hat{y}^{tidy}_{:t}$ $\gets$ [$\hat{y}^{tidy}_{:t-1}$, $\hat{y}_{i-1}$] \Comment{\scriptsize Replace current prediction with last segments \small}
                                                \State $\textbf{break}$
                                            \EndIndent
                                        \EndIndent \State else:  \Comment{\scriptsize If raw segment frame prediction reoccurs within b frames \small}
                                            \Indent \State $\textit{jump\_idx}$ $\gets$ \textbf{first\_index} where $\hat{y}_{i-b:i} == \hat{y}_{t}$
                                            \State for j in range($b-\textit{jump\_idx}-1)$:
                                            \Indent \State skip $i'th$ iteration step \Comment{\scriptsize Jump back to the beginning of the segment \small}
                                        \EndIndent
                                    \EndIndent
                                \EndIndent
                            \EndIndent
                        \EndIndent
                    \EndIndent
\end{algorithmic}
\end{algorithm}

When operating action recognition models on densely sampled input clips (with a fixed temporal stride $\tau$ between frames) in order to facilitate the learning of features across a full action segment during training the input clips initial frame is typically randomly sampled between $N_s$ and $N_e$ - $T\tau$, where $N_s$ and $N_e$ are the start and end frame index of the labelled segment within an extended video containing N frames. This traditional dense sampling strategy introduces two major issues when operating during online inference on short atomic actions common within HRI, which are not detected during offline pre-cropped segment level model evaluation. 

Firstly, as training clip creation boundaries are constricted to within the labelled segment, there is a T$\tau/2$ frame delay at the beginning and end of a segment before the online sampled clip will begin to look like an example seen by the model during training\footnote{This problem still arises when operating with a multi-frame fully causal model as the training models will not match inference clips until $T\tau$ frames have passed.} resulting in inaccurate segment boundaries, as well as models completely missing short atomic action segments. Secondly, when operating on very short ($N_e - N_s < T\tau$) actions, ``out of bound'' frame indices are generally replaced by either repeating the start and end frame or by wrapping the sampling indices to ensure they fall within the action segment boundaries. This training clip modification at segment boundaries introduces a significant discrepancy between AR based training and sliding window inference clips.

In order to combat these two issues with traditional dense sampling, we introduce surround dense sampling whereby the initial frame is selected as described in table \ref{tab:sample} and depicted in Fig. \ref{fig:overview}. Within surround sampling, the densely sampled frame indices of a training clip are allowed to extend to $T\tau/2$ frames beyond the boundaries of the ground truth segment label. This training sampling approach allows densely sampled online sliding window clips across a full segment to match offline training clips seen by the model during training.

\subsection{\textit{O-TALC} Algorithm} \label{sec:talc}

As we focus on online AS rather than action detection we introduce the simple yet surprisingly effective \textit{O-TALC} algorithm to combat oversegmentation in an online manner via the explicit real time removal of any short erroneous segments which fall below a pre-defined cut-off value.

Outlined in algorithm \ref{alg:TALC}, \textit{O-TALC} takes the most recently produced frame prediction $\hat{y}_{t}$, a history of previous raw frame predictions $\hat{y}_{:t-1}$ and current tidied predictions $\hat{y}^{tidy}_{:t-1}$ and produces a new set of tidied frame predictions up to the current frame, which can be passed to a downstream application requiring real time frame predictions. Shown graphically in Fig. \ref{fig:overview} the \textit{O-TALC} algorithm is designed to not classify incoming frames belonging to a new segment until the segment (with an additional over segmentation joining parameter, $b$) reaches a \texttt{full\_seg\_length} longer than a predefined cutoff length $C_{min}$, at which point the algorithm will backdate tidied output frame predictions to be passed to an external HRI framework.

\begin{table}[]
\resizebox{0.48\textwidth}{!}{\begin{tabular}{c|ccc|ccc} \toprule
\multicolumn{1}{c|}{\multirow{2}{*}{\textbf{\begin{tabular}[c]{@{}c@{}}Testing\\ Strategy\end{tabular}}}} & \multicolumn{3}{c|}{\textbf{Mean (dense/surround)}} & \multicolumn{3}{c}{\textbf{Min (dense/surround)}} \\
\multicolumn{1}{c|}{} & Precision & Recall & F$_1$ & Precision & Recall & F$_1$ \\ \hline
\begin{tabular}[c]{@{}c@{}}Pre-cropped \\ Segment\end{tabular} & 90.7/88.6 & 89.2/87.1 & 89.5/87.3 & 70.0/76.0 & 62.0/44.0 & 75.0/52.0 \\ \hline
\begin{tabular}[c]{@{}l@{}}Extended\\ Sequence\end{tabular} & 56.3/84.6 & 59.9/75.0 & 52.5/78.7 & 4.0/31.8 & 20.0/52.8 & 6.7/42.4 \\ \bottomrule
\end{tabular}}
\caption{\small Comparison in performance on a pre-cropped (AR) and extended AS (segment frame averaged predictions) between dense and surround sampling during training on the CBAA dataset.} \label{tab:dense_vs_surround}
\vspace{-10pt}
\end{table}

In order to identify the optimal cutoff point $C_{min}$, two approaches are investigated within this work:

\textbf{Static Cutoffs.} Assuming all erroneous short segment and oversegmentation errors fall below a given threshold a static class agnostic $C_{min}$ is adopted, with the optimal values determined through a hyper parameter search across validation data, with search values being informed by observing class statistics for a given dataset.

\textbf{Class Based}. Due to the large variation and very short class lengths present within many HRI AS datasets (such as CBAA), we propose an additional class based minimum cutoff length, $C^i_{min}$ calculated via $C^i_{min}$ = $\mu_{c_i}$ - $\kappa$*$\sigma_{c_i}$, where $\mu_{c_i}$ and $\sigma_{c_i}$ are the mean and standard deviations of a fitted log normal distribution to the ground truth training segments for class $i$. In practice to allow for inaccuracies in segment statistical modelling an additional minimum $C^{abs}_{min}$ is adopted. Optimal values of $\kappa$ and  $C^{abs}_{min}$ are found via a grid search on validation data.

It should be noted that while \textit{O-TALC} operates in real time it adds a small delay to the prediction of new segment of $C_{min}$ frames. As shown in the results, this cutoff is short; typically resulting in minimum of <1 second delay on the CBAA dataset, and a $\sim$3 seconds on 50salads and Assembly-101. Furthermore, when operating with class based cut-offs $C_{min}^i$ is dependent on the mean length of new action segment class, allowing short actions to be identified more quickly.

\begin{figure}
    \centering
    \includegraphics[width=0.48\textwidth]{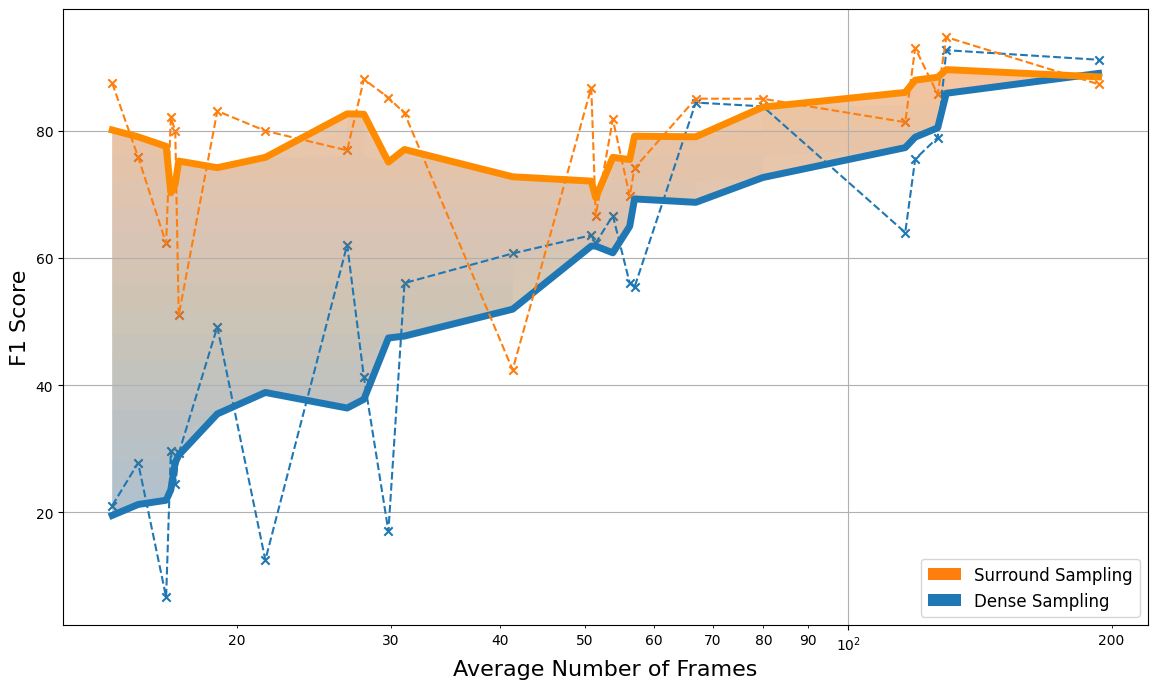}
    \caption{\textbf{F}$_1$ performance \textit{vs.} average class length on an extended sequence level for two models trained with dense and surround dense sampling on a real world CBAA dataset.} \label{fig:f1_vs_length}
    \vspace{-15pt}
\end{figure}

\begin{table}[]
\small
\resizebox{0.48\textwidth}{!}{\begin{tabular}{ll|l|l|lll|l}
\toprule
\multicolumn{2}{l|}{\begin{tabular}[c]{@{}l@{}} \textbf{Segmentation} \\ \textbf{Approach (Online?)} \end{tabular}} & \textbf{\begin{tabular}[c]{@{}l@{}}Sampling \\ Approach\end{tabular}} & \textbf{Acc} & \multicolumn{3}{l|}{\textbf{F1@\{0.5,0.25,0.1\}}} & \textbf{Edit} \\ \hline
\multicolumn{2}{l|}{\multirow{2}{*}{Sliding Window (\ding{51})}} & Dense  & 71.7 & 20.2 & 31.3 & 37.1 & 33.6 \\
\multicolumn{2}{l|}{} & Surround & 79.9 & 44.7 & 52.2 & 54.6 & 40.4 \\ \hline
\multicolumn{2}{l|}{\begin{tabular}[c]{@{}l@{}}Recursive \\ Averaging \cite{Efficient} (\ding{51}) \end{tabular}} & Surround & 79.9 & 52.9 & 64.8 & 64.8 & 52.1 \\
\multicolumn{2}{l|}{\begin{tabular}[c]{@{}l@{}}Modal Value \\ Smoothing \cite{Efficient} (\ding{51})\end{tabular}} & Surround & 80.0 & 66.3 & 79.6 & 81.4 & 73.0 \\ \hline
\multirow{2}{*}{\begin{tabular}[c]{@{}l@{}} Static \\ \textit{O-TALC} (\ding{51}) \end{tabular}} & {\scriptsize$C_{min}$,b =10,5} & Dense & 71.7 & 40.1 & 60.5 & 66.7 & 55.5 \\
 & {\scriptsize$C_{min}$,b = 9,2} & Surround & 80.3 & 72.0 & 83.9 & 85.7 & 78.1 \\
\multirow{2}{*}{\begin{tabular}[c]{@{}l@{}} Class Based \\ \textit{O-TALC} (\ding{51}) \end{tabular}} & {\scriptsize$\kappa$,$C_{min}^{abs}$,b = 2,4,3} & Dense & 71.9 & 42.3 & 61.8 & 67.7 & 56.5 \\
 & {\scriptsize$\kappa$,$C_{min}^{abs}$,b = 2,2,2} & Surround & 80.3 & 73.6 & 85.1 & \textbf{86.7} & \textbf{79.9} \\ \hline
\multicolumn{2}{l|}{\multirow{3}{*}{ms-TCN \cite{ms-TCN} (\xmark)}} & \begin{tabular}[c]{@{}l@{}}Kinetics \\ Features\end{tabular} & 70.7 & 51.2 & 68.5 & 73.8 & 71.4 \\
\multicolumn{2}{l|}{} & Dense & 82.5 & 72.1 & 83.4 & 85.4 & 78.8 \\
\multicolumn{2}{l|}{} & Surround & \textbf{84.1} & \textbf{77.6} & \textbf{85.3} & 86.3 & 79.6 \\ \hline
\end{tabular}}
\caption{Sequence level IoU results for several online segmentation approaches compared to the offline ms-TCN on the CBAA dataset.} \label{tab:CBAA_results}
\vspace{-12pt}
\end{table}

\section{Experiments}

\begin{table*}[]
\small
\begin{tabular}{c|l|c|c|c|ccc|c} \toprule
\textbf{Dataset} & \textbf{Method} & \textbf{Online} & \multicolumn{1}{c|}{\textbf{\begin{tabular}[c]{@{}c@{}}Sliding Window \\ Size (Frames)\end{tabular}}} & \multicolumn{1}{c|}{\textbf{Acc (MoF)}} & \multicolumn{3}{c|}{\textbf{F1@\{0.1,0.25,0.5\}}} & \multicolumn{1}{c}{\textbf{Edit}} \\ \hline
\multirow{8}{*}{50Salads \cite{50salads}} & ms-TCN \cite{ms-TCN} & \xmark & N/A & 80.7 & 76.3 & 74.0 & 64.5 & 67.9 \\
 & ms-TCN++ \cite{ms-tcn++} & \xmark & N/A & 83.7 & 80.6 & 78.7 & 70.1 & 74.3 \\
 & C2F-TCN \cite{C2F-TCN} & \xmark & N/A & 84.9 & 84.3 & 81.7 & 72.8 & 76.3 \\
 & L-Context \cite{long-context} & \xmark & N/A & 87.7 & 89.5 & 88.1 & 82.4 & 84.1 \\
 & DiffAct \cite{diff-AS} & \xmark & N/A & 88.9 & 90.1 & 89.2 & 83.7 & 85.0 \\ \cline{2-9} 
 & Transeger \cite{Transeger} & \cmark & 32 & 82.5 & 55.0 & - & - & - \\
 & Kang et al. \cite{Efficient} & \cmark & 15 & 78.7 & 70.1 & 66.3 & 60.8 & 65.8 \\
 & \textit{O-TALC} {\footnotesize(Static - $C_{min}$,b = 125,50)} & \cmark & 16 & 78.9 & 80.2 & 77.7 & 70.6 & 70.0 \\ 
 & \textit{O-TALC} {\footnotesize(Class Based - $\kappa,C_{min}^{abs}$,b = 1.5,10,15)} & \cmark & 16 & 78.3 & 79.9 & 77.4 & 67.0 & 70.6 \\ 
 \hline \hline
\multicolumn{1}{l|}{\multirow{5}{*}{Assembly-101 \cite{Assembly-101}}} & ms-TCN++ (from \cite{Assembly-101}) & \xmark & N/A & 37.1 & 31.6 & 27.8 & 20.6 & 30.7 \\
\multicolumn{1}{l|}{} & C2F-TCN (from \cite{Assembly-101}) & \xmark & N/A & 39.2 & 33.3 & 29.0 & 21.3 & 32.4 \\
\multicolumn{1}{l|}{} & L-Context* \cite{long-context} & \xmark & N/A & 41.2 & 33.9 & 30.0 & 22.6 & 30.4 \\ \cline{2-9} 
 & \textit{O-TALC}* {\footnotesize(Static - $C_{min}$,b = 110,5)} & \cmark & 8 & 34.6 & 33.5 & 30.1 & 21.7 & 26.2 \\
& \textit{O-TALC}* {\footnotesize(Class Based - $\kappa,C_{min}^{abs}$,b = 1.5,50,20)} & \cmark & 8 & 33.2 & 34.5 & 30.2 & 21.6 & 27.1 \\ \bottomrule
\end{tabular}
\caption{Comparison between state-of-the-art RGB based offline and online methods for AS on 50Salads and Assembly-101 datasets. *Evaluation is performed on the publicly available validation split of Assembly-101 dataset} \label{tab:sota_results}
\vspace{-5pt}
\end{table*}

To ensure the relevance of our contributions to real world human robot collaboration scenarios, we perform surround sampling ablation experimentation on the real world atomic action manufacturing CBAA dataset designed for assembly understanding outlined in \cite{hands} due to the rapid and short range nature of the actions present. We further compare our approach to state-of-the-art models on the challenging Assembly-101 \cite{Assembly-101} and longer segment 50 Salads \cite{50salads} datasets.

\textbf{Implementation Details.} In order to facilitate optimal performance on fine-grained actions which require short range spatio-temporal learning for classification \cite{hands}, we adopt a temporal segment network \cite{TSN} frame averaging approach for classification with inserted TSMs to facilitate spatio-temporal learning with hyper-parameters values matching \cite{TSM}.

A T=8,$\tau$=8 Resent-50 \cite{resnet} backbone is utilised for the CBAA and Assembly-101 dataset, with a T=16,$\tau$=6 MobileNet-v2 \cite{mobnet-v2} model utilised for 50 Salads to allow for a fair comparison to work by Kang et al. \cite{Efficient}. The CBAA and 50 Salad models are trained from Kinetics \cite{kinetics} pre-trained weights with a learning rate of 0.002 for 100 epochs, decreasing by a factor of 10 at 60 and 80 epochs, with a frozen batch normalisation parameters to reduce overfitting. The Assembly-101 model is trained only on view v-2 (above camera) in two fine-tuning stages, firstly from EPIC-KITCHEN \cite{epic-kitch} weights on the fine-grained action recognition labels for 50 epoch with a learning rate of 0.008, decreasing by a factor of 10 at 30 and 40 epochs, with a secondary fine-tuning stage on the coarse AS labels with a learning rate of 0.001 for 40 epochs, decreasing by a factor of 10 at 20 and 30 epochs.

In order to identify optimal \textit{O-TALC} hyper-parameters the following grid-search are implemented on validation data:

\begin{itemize}
    \item \textbf{CBAA:} \textbf{Static} $C_{min}$=\{2,3,4,5,6,7,8,9,10,12,15,20\},b=\{1,2,3,4,5,6\}. \newline \textbf{Class Based:} $\kappa$ = \{1,1.5,2,2.5,3\}, $C^{abs}_{min}$ = \{1,2,3,4,5\}, b selected from between 1 and $C^{abs}_{min}$ for each value of $C^{abs}_{min}$. 

    \item \textbf{50 Salads:} \textbf{Static} $C_{min}$=\{5,10,20,50,75,100,125,150,200,300\}, \newline
    b =\{1,3,5,10,15,20,50\}. \textbf{Class Based} $\kappa$ = \{1.5,2,2.5,3\}, \newline $C^{abs}_{min}$ = \{10,30,40\}, b=\{1,3,5,10,15,20\}. 

\item \textbf{Assembly-101: Static} $C_{min}$=\{20,50,70,90,110,150,200\}, \newline
b=\{1,2,5,10,20\}. \textbf{Class Based:} $\kappa$ = \{1,1.5,2,2.5\},
\newline $C^{abs}_{min}$ = \{20,30,40,50\}, b=\{2,5,10,20\}.

\end{itemize}

\textbf{Evaluation Metrics.} Following standard practice we report the frame level accuracy, $F_1$ score at an IoU threshold overlap of 0.5, 0.25, 0.1 \cite{TCN} and the edit score \cite{edit_score} on an extended sequence length. For sequence level class analysis we report frame averaged predictions over ground truth segment boundaries.

\subsection{Importance of Surround Sampling}

Table \ref{tab:dense_vs_surround} highlights the significant model degradation observed when operating in an online a sliding window fashion when training with the traditional densely sampled frame paradigm. Whilst surround sampling is observed to harm the model performance on a pre-cropped segment level evaluation (akin to action recognition), densely sampled models perform extremely poorly on extended sequences, dropping from a mean F$_1$ of 89.5$\rightarrow$52.5, with surround sampled models seeing a more modest drop from 87.3$\rightarrow$78.7. This observed drop in performance is not shared evenly across classes with Fig. \ref{fig:f1_vs_length} showing the majority of the performance degradation on a sequence level is due to the poor performance on shorter actions due to the increasing discrepancy between training and online inference clips. This degradation is observed to accelerate below a length of $\sim$60 frames, corresponding to the length of training clips (T*$\tau$=8*8=64), below which index truncation or the wrapping of frames become prevalent within training clips, resulting in the worse performing class seeing an $F_1$ score drop from 75.0$\rightarrow$6.7.

\begin{figure}[]
    \centering
    \includegraphics[width=0.5\textwidth]{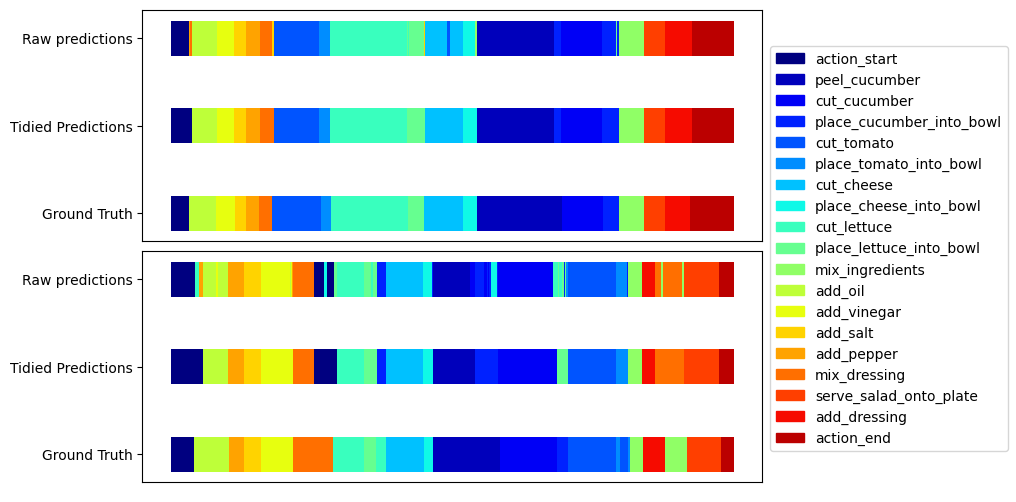}
    \caption{Example of \textit{O-TALC} frame tidying on 50Salads split 1 (upper) and split 3 (lower) test sequences.}
    \label{fig:pred_example}
    \vspace{-10pt}
\end{figure}

Table \ref{tab:CBAA_results} demonstrates the performance of dense \textit{vs.} surround training sampling on IoU threshold based metrics for a variety of segmentation approaches. Surround sampling is found to continuously improve all segmentation approaches, and even leads to improvements when operating offline segmentation approaches, such as ms-TCN. Whilst the improvement is more modest (F$_1$@0.5 72.1$\rightarrow$77.6) compared to improvements on class based \textit{O-TALC} ($F_1$@0.5 42.3$\rightarrow$ 73.6) it highlights that surround sampling can be utilised to improve the base feature extraction step for more complex two stage offline segmentation approaches.

\subsection{\textit{O-TALC} Performance}

Table \ref{tab:CBAA_results} further compares the performance of \textit{O-TALC} operating with optimal static and class based cutoff values to label cleaning strategies outlined by Kang et al. \cite{Efficient}, on the CBAA dataset. Both static and class based cutoff values are observed to significantly outperforms the recursive averaging or modal value smoothing of softmax features. Due to \textit{O-TALC}s ability to explicitly remove short segments, an increased precision (decrease in the number of false positive oversegmentation predictions) is observed as the driving force of this improvement, while segment recall remains consistent. Furthermore, it is observed that when operating on short actions such as those in CBAA, class based cutoffs lead to improved segmentation results, due to the ability to retain more of the shortest action classes which have average lengths <15 frames, similar to the optimal static $C_{min}$ = 10.

Table \ref{tab:sota_results} compares a surround trained sliding window model operating with both static and class based \textit{O-TALC} to a representative sample of offline as well as online AS approaches on the 50 Salads and Assembly-101 dataset. Both the static and class based cutoff values outperform the online AS results reported by Kang \textit{et al.} \cite{Efficient} on the 50 Salads dataset. In contrast to CBAA dataset, static cutoff value are found to outperform class based cutoffs on 50 Salads, potentially due to the much longer average class lengths, with the minimum class average length being 230 frames; comfortably above the optimal static $C_{min}$ = 125. 

Finally, we provide the first online AS results on the Assembly-101 dataset, outperforming ms-TCN++ \cite{ms-tcn++} and matching the performance of other state-of-the-art offline approaches when considering $F_1$@IoU metrics. Whilst longer term temporal modelling is undoubtedly important for accurate segmentation (as seen by \textit{O-TALC}s relatively poor performance compared to offline fully temporally aware approaches on 50 Salads), due to the extremely challenging nature of the Assembly-101 dataset much of the performance gain seen by \textit{O-TALC} is through the explicit removal of the vast majority of short incorrect segment predictions, rather than more accurate ground truth segment predictions (as observed via the lower Edit score). This is further supported within Fig. \ref{fig:pred_example} which highlights how \textit{O-TALC} can reduce the number of inaccurate segments, however generally does not improve the overall segment predictions, allowing additional performance gains to be achieved through stronger backbone model predictions.

\vspace{-5pt}
\section{Conclusion}

In this short paper we introduce surround dense sampling and \textbf{O}nline \textbf{T}emporally \textbf{A}ware \textbf{L}abel \textbf{C}leaning (\textit{O-TALC}) to facilitate improved segment boundary predictions and reduce oversegmentation when performing online AS. As our approaches are invariant to model backbone and we opt to implement short range spatio-temporal models in simple sliding window fashion with no long term modelling, we believe additional longer term causal temporal modelling, often adopted in action detection e.g. \cite{long-transformer, distilation-OAD} could be incorporated for improved segment classification within future work. Alternatively, if low latency is a priority, buffer style networks\cite{TSM, movienet} operating on single input frames with access to previous frame features could be adopted alongside surround sampling and \textit{O-TALC} to provide reduced online segmentation latency.

\bibliographystyle{ACM-Reference-Format}
\bibliography{sample-base}

\end{document}